\documentclass[conference]{IEEEtran}
\IEEEoverridecommandlockouts
\usepackage{cite}
\usepackage{amsmath,amssymb,amsfonts}
\usepackage{algorithmic}
\usepackage{graphicx}
\usepackage{textcomp}
\usepackage{xcolor}
\usepackage[colorlinks]{hyperref}
\usepackage{soul, color}

\def\BibTeX{{\rm B\kern-.05em{\sc i\kern-.025em b}\kern-.08em
    T\kern-.1667em\lower.7ex\hbox{E}\kern-.125emX}}
\begin{document}

\title{\textbf{skrl}: Modular and Flexible Library for Reinforcement Learning\\
\thanks{This study was partially financed by H2020-WIDE SPREAD project no. 857061 ``Networking for Research and Development of Human Interactive and Sensitive Robotics Taking Advantage of Additive Manufacturing -- R2P2'' and the H2020-ECSEL JU project no. 876852 ``Verification and Validation of Automated Systems' Safety and Security - VALU3S'''.}
}

\makeatletter 
\newcommand{\linebreakand}{%
  \end{@IEEEauthorhalign}
  \hfill\mbox{}\par
  \mbox{}\hfill\begin{@IEEEauthorhalign}
}
\makeatother 

\author{\IEEEauthorblockN{1\textsuperscript{st} Antonio Serrano-Muñoz}
\IEEEauthorblockA{\textit{Dept. of Robotics and Automation} \\
\textit{Mondragon Unibertsitatea}\\
Arrasate, Spain \\
aserrano@mondragon.edu}
\and
\IEEEauthorblockN{2\textsuperscript{nd} Dimitris Chrysostomou}
\IEEEauthorblockA{\textit{Dept. of Materials and Production} \\
\textit{Aalborg University}\\
Aalborg, Denmark \\
dimi@mp.aau.dk}
\and
\IEEEauthorblockN{3\textsuperscript{rd} Simon B{\o}gh}
\IEEEauthorblockA{\textit{Dept. of Materials and Production} \\
\textit{Aalborg University}\\
Aalborg, Denmark \\
sb@mp.aau.dk}
\linebreakand
\IEEEauthorblockN{4\textsuperscript{th} Nestor Arana-Arexolaleiba}
\IEEEauthorblockA{\textit{Dept. of Robotics and Automation} \\
\textit{Mondragon Unibertsitatea / Aalborg University}\\
Arrasate, Spain \\
narana@mondragon.edu}
}

\maketitle

\begin{abstract}

skrl is an open-source modular library for reinforcement learning written in Python and designed with a focus on readability, simplicity, and transparency of algorithm implementations. In addition to supporting environments that use the traditional interfaces from OpenAI Gym and DeepMind, it provides the facility to load, configure, and operate NVIDIA Isaac Gym and NVIDIA Omniverse Isaac Gym environments. Furthermore, it enables the simultaneous training of several agents with customizable scopes (subsets of environments among all available ones), which may or may not share resources, in the same run. The library's documentation can be found at \url{https://skrl.readthedocs.io} and its source code is available on GitHub at \url{https://github.com/Toni-SM/skrl}.

\end{abstract}


\section{Introduction}
As a Machine Learning subfield, Reinforcement Learning (RL) is a paradigm to learn, improve and generalize the decision-making capabilities of autonomous agents. RL allows agents to learn through interaction with their environments and, ideally, generalize the learned behavior to new, unseen scenarios. 

As shown in \autoref{fig:libraries}, there has been an increase in the development of RL libraries for research and applications in recent years. This explains well the increase in popularity in the research community from breakthroughs in RL around 2014-2015, but also the boost in successful real-world applications and access to efficient simulation tools and deep learning frameworks. 


\begin{figure}[htbp]
    \centerline{\includegraphics[width=0.48\textwidth]{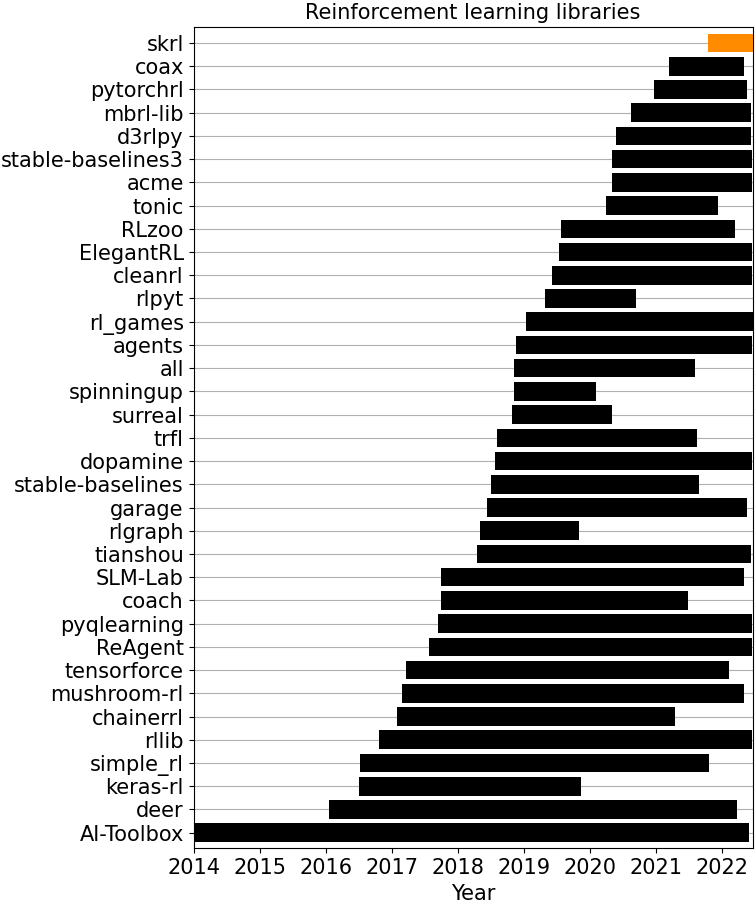}}
    \caption{Main RL libraries' lifecycle. The lifecycle is computed using the repository's creation date and the last commit message retrieved from GitHub.}
    \label{fig:libraries}
\end{figure}

Three fundamental milestones mark the rise of RL in our times. 1) The development of new learning algorithms, especially those that use artificial neural networks as approximation functions (Deep RL). 2) The development of Gym by OpenAI (2016). It exposes a common interface for designing and standardizing environments~\cite{brockman2016openai}. 3) The development of benchmarking scenarios in areas such as video games and gaming, autonomous navigation, and robotics. Those benchmarks have been widely accepted by the scientific community allowing to compare results between different implementations. 

Particularly in robotics and autonomous systems, physics-based simulators play an essential role. Simulation enables better time management, cost reduction, and safety in safety-critical and/or complex settings, especially during exploration~\cite{korber2021comparing}.

MuJoCo~\cite{todorov2012mujoco} and PyBullet~\cite{coumans2021} are physics engines that facilitate research and development in e.g. robotics. They provide fast and accurate simulation for rigid multi-body dynamics and control. Such simulation tools make it possible to scale up training for advanced contact-rich environments. OpenAI Gym~\cite{brockman2016openai} and DeepMind Environment~\cite{dm_env2019}\cite{tunyasuvunakool2020dm_control} are open source interfaces for RL tasks. They provide a suite of tasks for getting started with RL. Besides, they define a standard interface between the agent and the environment. Specifically, in OpenAI Gym, it is composed of \textit{actions} (to be sent to the environment), and \textit{observation}, \textit{reward}, and whether an episode is \textit{done} (received from the environment). This interface definition has become standard in RL research and development.

With the release of Isaac Gym Preview, and recently Omniverse Isaac Gym, a GPU-based physics simulation platform from NVIDIA, a new generation of robotic simulation with tens of thousands of simultaneous environments on a single GPU has emerged~\cite{makoviychuk2021isaac}. They allow researchers to easily run massive experiments using an OpenAI Gym-like API by offloading both physics simulation and neural network training onto the GPU. While Isaac Gym and Omniverse Isaac Gym provide some examples for modeling the environment, a streamlined interface towards implementing RL algorithms in a flexible and modular way is needed.

In this work, we present \textbf{skrl}, an RL library designed with the following principles in mind: 1) modularity, leaving room for each component to be interchangeable and making it possible to create more complex systems. 2) readability, simplicity, and transparency of the algorithm implementations, which reduces the learning curve with an educational approach. 3) support for different interfaces and 4) simultaneous learning on Isaac Gym and Omniverse Isaac Gym.

The rest of this document is organized as follows. Related works are analyzed in \autoref{sec:related_works}. The description of the implementation and features are presented in \autoref{sec:features}. An evaluation and comparison of experiments as a performance measure are discussed in \autoref{sec:evaluation} while we conclude the paper in \autoref{sec:conclusion}.

\section{Related work}
\label{sec:related_works}

Although there are significant differences among all the RL libraries shown in \autoref{fig:libraries}, some of them share common features with the proposed library.

Modularity is a desirable feature for the scalability and flexibility of a system and the reusability of its constituent components. ChainerRL~\cite{fujita2021chainerrl}, a library built on top of Chainer, and  PyTorchRL~\cite{bou2020pytorchrl} are developed around the idea of agent composability. They provide a set of building blocks for the development of new agents. rlpyt~\cite{stooke2019rlpyt}, Tonic~\cite{pardo2020tonic}, and MushroomRL~\cite{d2020mushroomrl} also offer building blocks as configurable modules, but their designs are based on a hierarchy of inheritances involving many files and lack consistent naming in various implementations.

The code's readability, simplicity and transparency are indispensable for understanding implementations and using existing code or APIs to develop new RL methods; even more when small implementation details can significantly affect the performance of the algorithms~\cite{engstrom2019implementation}. Many libraries encapsulate great features deep in their coding, leading to difficulties in reproducibility such as RLlib~\cite{pmlr-v80-liang18b} or RLzoo~\cite{ding2020rlzoo}. Nevertheless, there are efforts in favor of readability, simplicity and transparency.

Spinning Up~\cite{SpinningUp2018}, from OpenAI, was implemented with an educational approach and detailed documentation. Stable Baselines3~\cite{StableBaselines3} offers readability and simplicity over modularity, focusing on model-free, single-agent algorithms. CleanRL~\cite{huang2021cleanrl} includes all the details of the algorithm and environment in a single file, arguing that it helps researchers understand the implementation and prototype new features. Although such compact implementation facilitates the setup of simple applications, library maintenance and addition of new features remain challenging.

Almost all RL libraries support the OpenAI Gym interface for learning environments. However, the same cannot be said for DeepMind Environment, Isaac Gym and Omniverse Isaac Gym. As mentioned in the introduction, the last two are recent and have a slightly different interface with OpenAI Gym.

In Isaac Gym's latest releases (preview 3 and 4) and Omniverse Isaac Gym, RL Games \cite{rl_games} is presented as the default library to run the example environments. ElegantRl~\cite{liu2021elegantrl} offers support for Isaac Gym environments. However, it only allows working with the previous release (preview 2), since it explicitly includes, within its source code, the original files of that preview.

Parallel learning attempts to increase the variety of data collection and/or the stability of the learning process. RLlib makes copies of the environments to scale experience collection for one worker or many workers on a single process or multiple processes on top of Ray. Although, its implementation, designed to provide a high-level API, makes it difficult to understand the code and perform custom experimentation. ElegantRl exploits the parallelism of RL algorithms at multiple levels. However, as mentioned above, it only supports the Isaac Gym (preview 2), and its parallelism at the worker and learner level (generating batches of actions and returning a transition batch) for vectorized environments only supports one agent.

\section{Implementation and features}
\label{sec:features}

skrl is an open-source modular library for RL written in Python (using PyTorch ~\cite{PyTorch}) and designed with a focus on readability, simplicity, and transparency of algorithm implementation. In addition to supporting the OpenAI Gym and DeepMind interfaces, it allows loading and configuring NVIDIA Isaac Gym and NVIDIA Omniverse Isaac Gym environments, enabling agents' simultaneous training by scopes (subsets of environments among all available environments), which may or may not share resources, in the same run. The following subsections describe its implementation and its main features.

\subsection{Structure and design concepts}

The file system structure that conforms the library is designed to group the components, according to their functionality, without mixing them. This design, focused on modularity, allows a quick understanding and use of the components by the researchers. 

As shown in \autoref{fig:schema}, the library is organized into seven components (and some utilities). The current implementation of the components is done using PyTorch~\cite{PyTorch}. However, the design of the file system allows for future implementations using other deep learning libraries such as TensorFlow~\cite{abadi2016tensorflow} or Chainer~\cite{tokui2015chainer}.


\begin{figure}[htbp]
    \centerline{\includegraphics[width=0.48\textwidth]{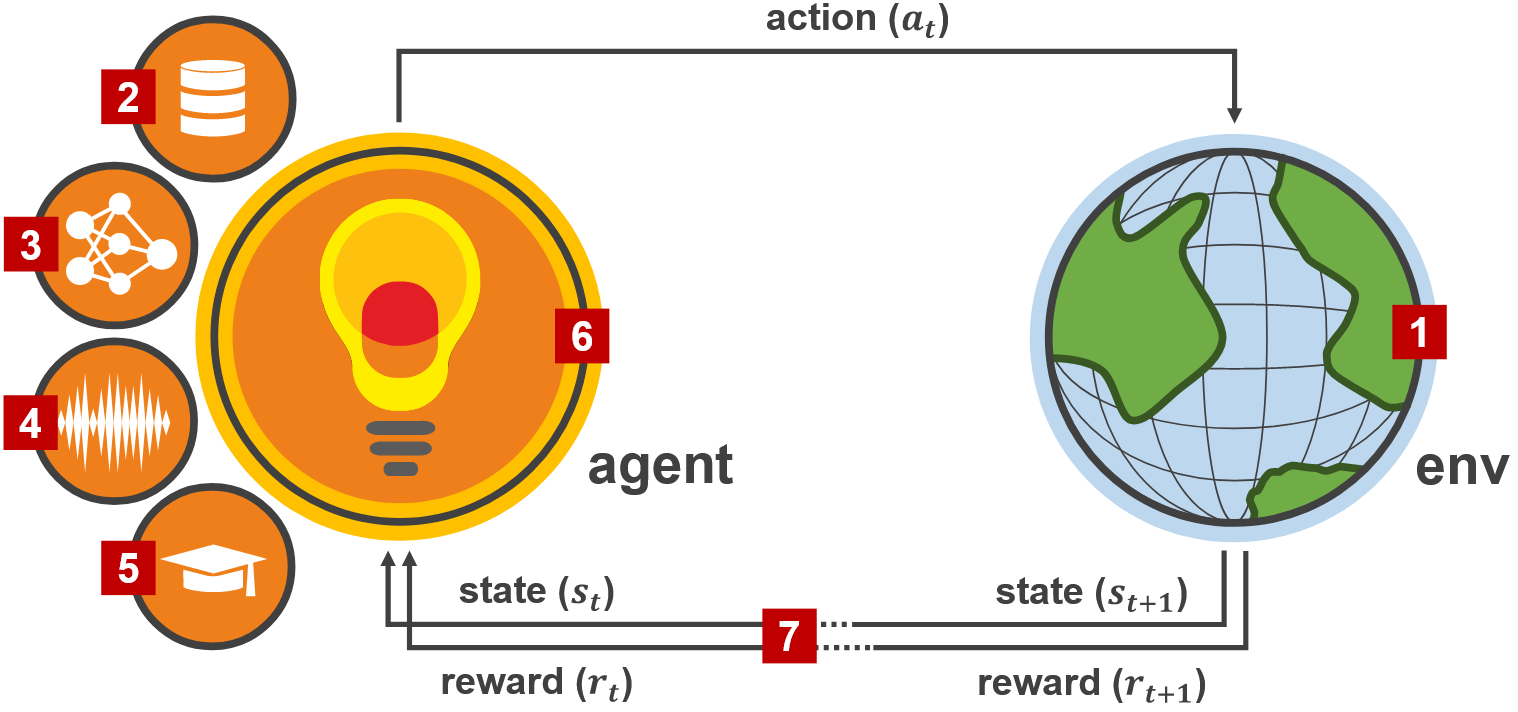}}
    \caption{Library components as part of the reinforcement learning schema: 1) environments, 2) memories, 3) models, 4) noises, 5) learning rate schedulers, 6) agents and 7) trainers.}
    \label{fig:schema}
\end{figure}


Except for the environments (\emph{envs}), all other components inherit properties and methods from one (and only one) base class implemented in a common file (\emph{base.py}) for each group. Apart from providing a uniform interface, the base classes implement common functionalities (which are not tied to the implementation details of the algorithms), such as logging to TensorBoard~\cite{abadi2016tensorflow} or saving and loading files to and from persistent storage. Focused on readability, simplicity, and transparency, each implementation within the same component is done standalone, even when two or more implementations may contain code in common.

The components that are part of skrl are as follows:

\begin{enumerate}
    \item \textbf{envs}: Definition of the loaders for Isaac Gym (preview 2, 3 and 4) and Omniverse Isaac Gym environments. It also includes the wrappers for each supported environment type: OpenAI Gym, DeepMind, Isaac Gym and Omniverse Isaac Gym.
    \item \textbf{memories}: Definition of generic memories. The memories are not bound to any agent (agents must create the internal tensors according to their specifications) and the implementations can be used as rollout buffer or experience replay memory, for example.
    \item \textbf{models}: Definition of helpers for building tabular models and function approximators using artificial neural networks. In contrast to other libraries, and to put the RL system’s control in the researchers’ hands, skrl does not provide policy definitions (this practice typically hides and reduces the system's flexibility, forcing developers to deeply inspect the code to make changes). Helper classes are provided to create discrete and continuous (stochastic or deterministic) policies within this component. In this case, the researcher is only concerned with the definition of artificial neural networks.
    \item \textbf{noises}: Definition of noises used by deterministic agents during the  exploration stage.
    \item \textbf{learning rate schedulers}: Definition of customized learning rate schedulers to adjust the learning rate of the optimizer between gradient steps or training epochs.
    \item \textbf{agents}: Definition of the RL methods that compute an optimal policy. The learning and optimization algorithm is implemented within a function under the same name (\emph{\_update}) in all cases. The following state-of-the-art methods are currently included as of this writing: CEM~\cite{szita2006learning},  DDPG~\cite{lillicrap2015continuous}, DQN~\cite{mnih2013playing}, DDQN~\cite{van2016deep}, PPO~\cite{schulman2017proximal}, Q-learning~\cite{watkins1989learning}, SAC~\cite{haarnoja2018soft}, SARSA~\cite{rummery1994line}, TD3~\cite{fujimoto2018addressing} and TRPO~\cite{schulman2015trust}. 
    \item \textbf{trainers}: Definition of the classes responsible for managing the agent's training and interaction with the environment. These definitions also allow the execution of simultaneous synchronous learning in Isaac Gym and Omniverse Isaac Gym.
\end{enumerate}

In addition, as mentioned above, a set of utilities are offered to perform, among others, the following operations: loading and post-processing of exported memory files and TensorBoard files, fast model instantiators, visualization of the environment's configuration and computation of inverse kinematics for robotic manipulators in Isaac Gym and Omniverse Isaac Gym.

\subsection{Support for different environment interfaces}

In order to work with a common logic and interface, and support interoperability between implementations, the trainers operate on wrapped environments. These wrappers allow experiments to be conducted in OpenAI Gym and DeepMind-like environments. The wrapped environment interface is based on the OpenAI Gym interface, as shown in \autoref{fig:wrapping}. 

\begin{figure}[htbp]
    \centerline{\includegraphics[width=0.495\textwidth]{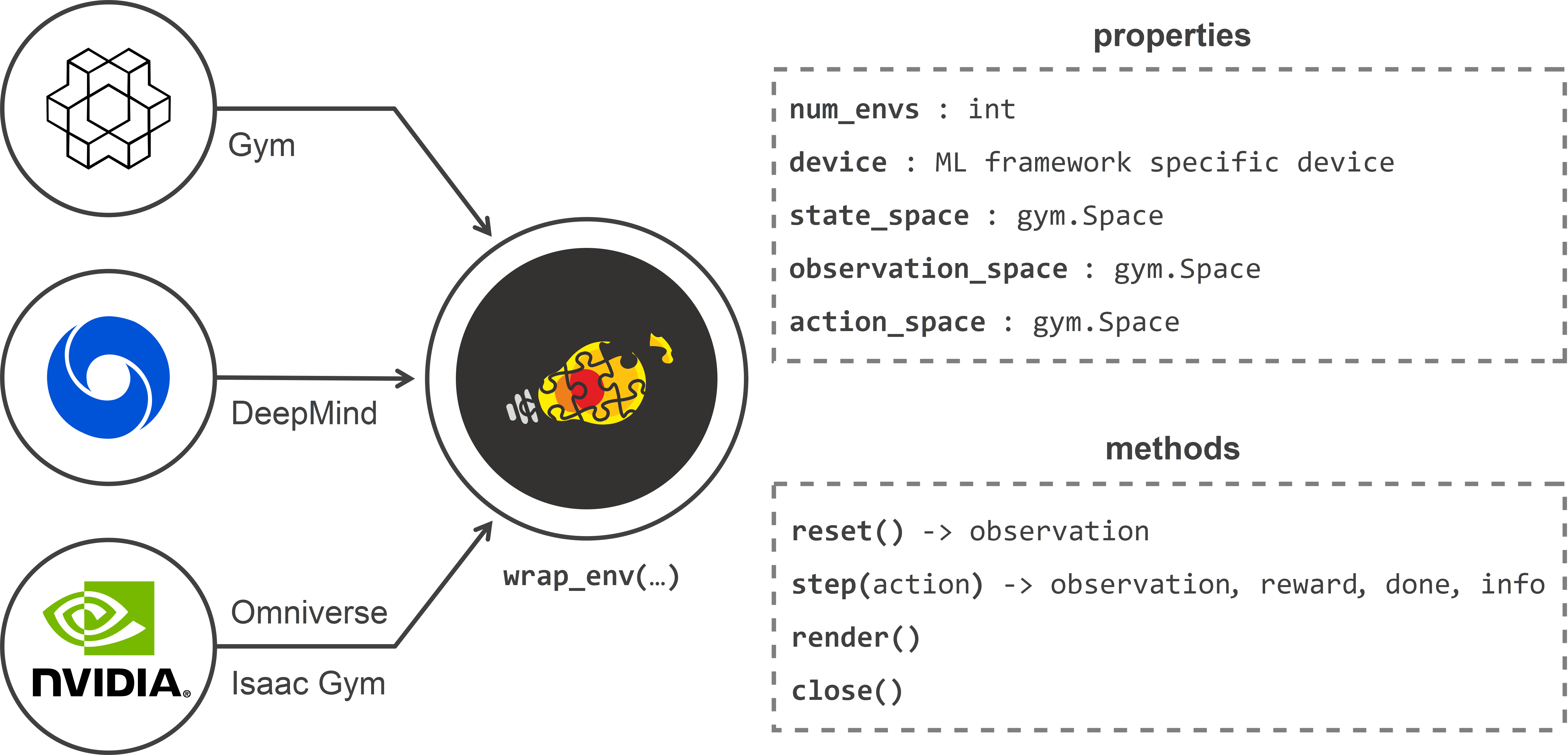}}
    \caption{Wrapped environment interface based on the OpenAI Gym interface.}
    \label{fig:wrapping}
\end{figure}

Two new properties are added: \emph{num\_envs}, which contains the number of parallel environments loaded, and \emph{device}, which stores information about the physical device (CPU or GPU) on which the simulation is running. In addition, this library enables the loading and configuration of Isaac Gym and Omniverse Isaac Gym environments by calling a single function. This function can handle the settings from command line arguments or from its parameters, as a python dictionary.

\subsection{Simultaneous learning by scopes in Isaac Gym and Omniverse Isaac Gym}

Isaac Gym (preview 2, 3 and 4) and Omniverse Isaac Gym simulate thousands of environments simultaneously by offering an API based on the vectorization of observations and actions. This library takes advantage of such parallelization by enabling the training of simultaneous agents (of the same or different classes) which may or may not share resources.

Each agent can define a working scope: a set of sub-environments among all available environments. Then, at each time step, the trainer collects the actions of each agent in their respective scopes and builds a single vector of actions that is passed to the physical simulation pipeline. After simulating the physics, the current state of observations, rewards and completed episodes are partitioned and passed to each agent, according to its scope, to execute the learning and optimization stage.

This setup makes it possible to compare, in a single run, the performance of several agents, hyperparameters and other components. Nevertheless, given this library's modular and flexible design, it also enables sharing resources between the different agents (such as the memory, for example) that can help improve the learning process.

\subsection{Documentation}

The documentation is written using reStructuredText and hosted online by Read the Docs under the url \url{https://skrl.readthedocs.io}. Apart from the library installation steps and API details (classes, functions, parameters and return values, etc.), snippets are also included to guide the development of new components or algorithms. In addition, a detailed mathematical description of the implementation of the RL agents is provided. Examples of use cases are included with their respective scripts and description of functionalities such as tracking and visualizing metrics.

\section{Evaluation}
\label{sec:evaluation}

A small subset of experiments\footnote{\textbf{Details and codes} for the experiments described in this section, and other experiments for OpenAI Gym, DeepMind, Isaac Gym and Omniverse Isaac Gym, can be found on the \textbf{documentation web page under \emph{Examples}: \url{https://skrl.readthedocs.io/en/latest/intro/examples.html}}.} have been performed in order to evaluate and compare the implementations of the algorithms with other RL libraries, and to exemplify the capability of working with OpenAI Gym, Isaac Gym (in its last three versions) and Omniverse Isaac Gym environments as shown in the \autoref{fig:envs}. For all case families, the same hyperparameter sets were used as far as the implementations of the involved RL libraries allowed.

\begin{figure}[htbp]
    \centerline{\includegraphics[width=0.48\textwidth]{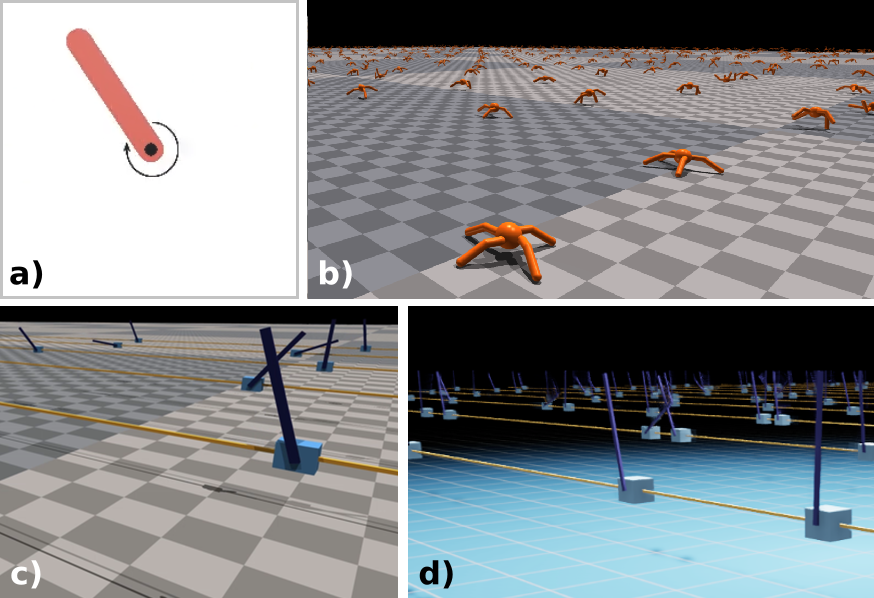}}
    \caption{Evaluation environment subset. a) Inverted pendulum (Pendulum-v0): OpenAI Gym classic control environment, b) Ant: NVIDIA Isaac Gym (preview 2), c) Cartpole: Isaac Gym (preview 3 and 4) and d) Cartpole: NVIDIA Omniverse Isaac Gym.}
    \label{fig:envs}
\end{figure}

The evaluations of the OpenAI Gym scenarios were performed in a docker container on a computer with a 2.20GHz Intel Xeon Silver 4114 CPU, 126GB of RAM and a NVIDIA RTX 2080Ti GPU. The Isaac Gym (preview 2, 3 and 4) and Omniverse Isaac Gym evaluations were performed on a workstation with a 3.00GHz Intel Xeon W-2295 CPU, 126GB of RAM and a NVIDIA RTX 6000 GPU.

\subsection{Learning in an environment with the OpenAI Gym interface}

\autoref{fig:reward-pendulum} shows the mean total reward and its standard deviation for the DDPG, TD3 and SAC agents of the skrl (ours), stable-baselines3 and RLlib libraries for the inverted pendulum environment. 

\begin{figure}[htbp]
    \centerline{\includegraphics[width=0.48\textwidth]{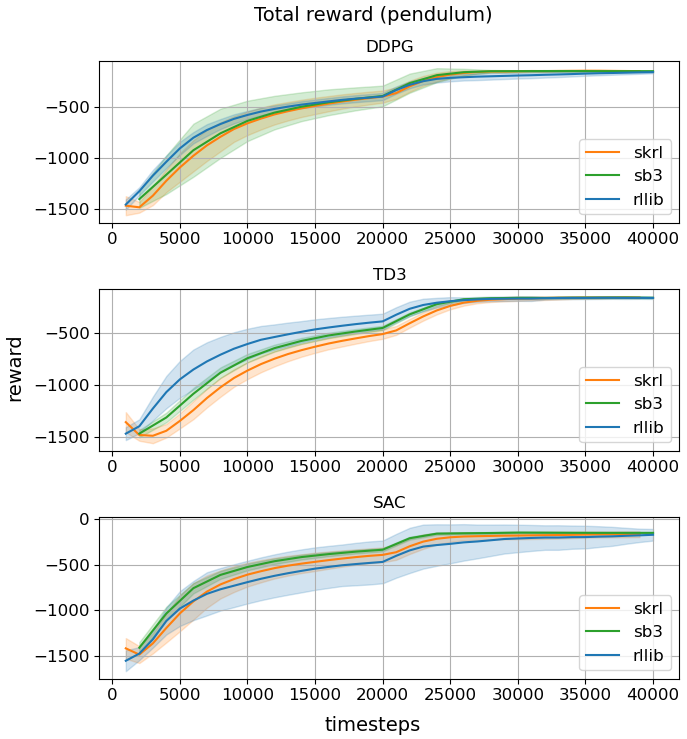}}
    \caption{Comparison of the total reward (mean and standard deviation) during training in the OpenAI Gym inverted pendulum environment for the skrl(ours), stable-baselines3 and RLlib libraries.}
    \label{fig:reward-pendulum}
\end{figure}

Although the different agents of the involved libraries have similar behavior in all cases, there are differences in training times. The execution of the task yielded comparable mean times (timesteps per second) for stable-baselines3 (DDPG: 140, TD3: 145, SAC: 77) and our library (DDPG: 135, TD3: 145, SAC: 70). The training times for RLlib, configured with a single worker, were three times slower than the results for the other libraries (DDPG: 44, TD3: 39, SAC: 22).

\subsection{Learning in Isaac Gym and Omniverse Isaac Gym environments}

\autoref{fig:reward-cartpole-ant} shows the mean total reward and its standard deviation for the PPO agents of the skrl (ours), rl\_pytorch (a small library included with the distribution of Isaac Gym preview 2) and rl\_games libraries for Cartpole and Ant environments. In this case, our library was evaluated in all Isaac Gym previews. 

\begin{figure}[htbp]
    \centerline{\includegraphics[width=0.48\textwidth]{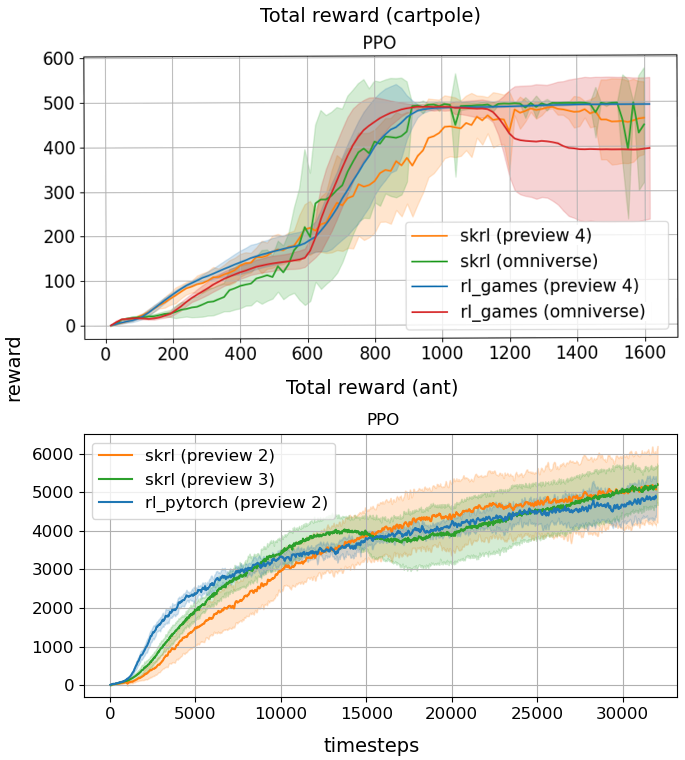}}
    \caption{Comparison of the total reward (mean and standard deviation) during training in the Cartpole (Isaac Gym preview 4 and Omniverse Isaac Gym) and Ant environments (Isaac Gym preview 2 and 3) for the skrl(ours), rl\_pytorch and rl\_games libraries.}
    \label{fig:reward-cartpole-ant}
\end{figure}

The libraries achieved comparative performance based on total reward and training time in all cases. The Cartpole averaged 105 timesteps per second for 512 environments, while the Ant averaged 45 timesteps per second for 1024 environments.

\subsection{Simultaneous learning by scopes in Isaac Gym}

\autoref{fig:parallel} illustrates an RL configuration in which three agents are trained in parallel by scopes. Each agent only interacts with a specific number of environments (DDPG is controlling 170 environments, TD3 is controlling 170 environments, SAC is controlling 172 environments) out of the entire set of available environments (512 environments). 

\begin{figure}[htbp]
    \centerline{\includegraphics[width=0.48\textwidth]{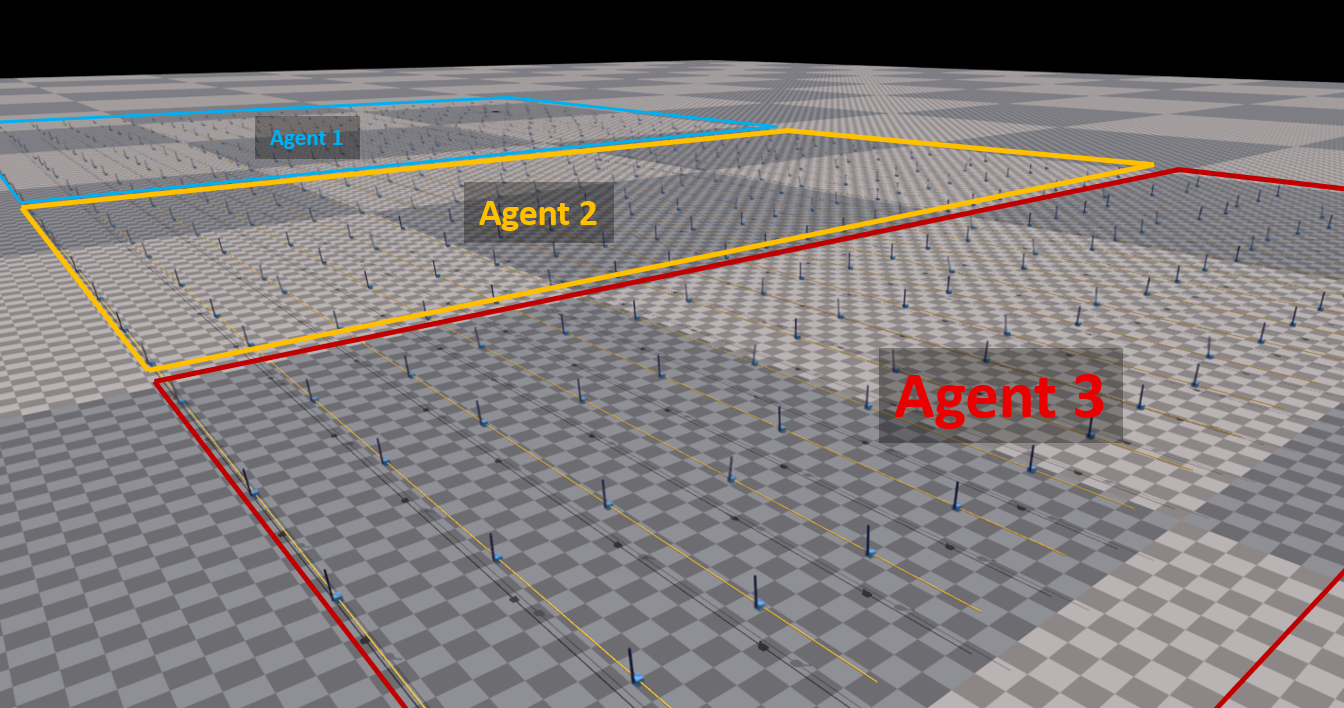}}
    \caption{Example of a simultaneous learning configuration in an Isaac Gym environment. The number of parallel environments is divided according to each agent scope.}
    \label{fig:parallel}
\end{figure}

For this configuration, \autoref{fig:reward-cartpole-parallel} shows the mean total reward and its standard deviation for two scenarios: simultaneous training without memory sharing and simultaneous training with memory sharing between the three agents.

\begin{figure}[htbp]
    \centerline{\includegraphics[width=0.48\textwidth]{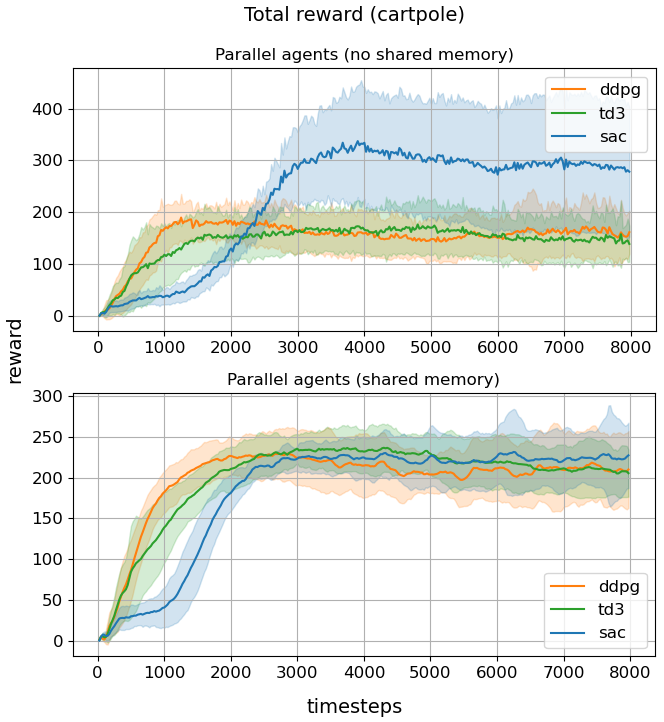}}
    \caption{Comparison of the total reward (mean and standard deviation) during training for the Cartpole environment (Isaac Gym preview 3). Top chart: standalone agents. Bottom chart: agents training in parallel sharing memory.}
    \label{fig:reward-cartpole-parallel}
\end{figure}

Even though the experiment was performed with a default and unoptimized hyperparameter set, there is a performance difference between using memory sharing and not. In the latter, a better and balanced performance is achieved.

\section{Conclusion}
\label{sec:conclusion}

skrl is a library for reinforcement learning that allows researchers to compose their experiments using a modular API. Its development has focused on readability, simplicity, and transparency of algorithm implementations, making it possible to reduce the learning curve's complexity and adaptations to the code. In addition, it supports training in environments with OpenAI Gym, DeepMind, Isaac Gym and Omniverse Isaac Gym interfaces.

Future work will include the implementation of other algorithms and components.

\section*{Acknowledgment}

We would like to express our gratitude for the funding and support received from NVIDIA under a collaboration agreement with the Mondragon Unibertsitatea.

\bibliographystyle{IEEEtran}
\bibliography{reference.bib}


\end{document}